# Enhancing GeoAI and location encoding with spatial point pattern statistics

A Case Study of Terrain Feature Classification

Sizhe Wang
School of Computing and
Augmented Intelligence
Arizona State University
Tempe, AZ, USA
wsizhe@asu.edu

Wenwen Li[†]
School of Geographical Sciences
and Urban Planning
Arizona State University
Tempe, AZ, USA
wenwen@asu.edu

## ABSTRACT

This study introduces a novel approach to terrain feature classification by incorporating spatial point pattern statistics into deep learning models. Inspired by the concept of location encoding, which aims to capture location characteristics to enhance GeoAI decision-making capabilities, we improve the GeoAI model by a knowledge driven approach to integrate both first-order and second-order effects of point patterns. This paper investigates how these spatial contexts impact the accuracy of terrain feature predictions. The results show that incorporating spatial point pattern statistics notably enhances model performance by leveraging different representations of spatial relationships.

## 1 INTRODUCTION

Geographic Artificial Intelligence (GeoAI) is an emerging field that leverages advanced computational techniques to analyze and interpret spatial data [1, 2]. One building block of the development of GeoAI research is the process of location encoding, which transforms geographic coordinates—such as latitude and longitude—into dense, continuous vector representations. This transformation enables deep learning models to effectively capture and utilize complex spatial relationships and patterns. As outline in this comprehensive review [3], the primary motivation behind location encoding is the inherent challenge of handling vector data directly within deep neural networks. The encoding process facilitates the integration of spatial context into these models, paving the way for more complex spatial analysis and application of geographic information.

Location encoding methods can be broadly categorized into two main techniques: direct learning and contrastive learning. Direct learning approaches involve training models using location coordinates as conditional inputs to enhance downstream tasks. Notable examples include the works of [4] and [5], which utilize the spatial context as geographic priors and loss function components respectively to refine the learning tasks. On the other hand, the works such as [6] and [7], encode the spatial information into semantic embeddings, providing more flexibility of incorporating such information into the deep learning models. Despite their effectiveness, these methods often fall short in addressing the complex interactions between locations and their intrinsic spatial properties.

In contrast, contrastive learning techniques have emerged as a powerful alternative, focusing on learning discriminative location embeddings by maximizing the similarity between spatial and imagery data. Pioneering studies, such as [8], [9], and [10], have demonstrated the potential of contrastive learning for generating robust location embeddings. However, these methods generally require extensive data sources, pretraining processes, and substantial computational resources. They also tend to emphasize the alignment between locations and imagery rather than capturing the detailed characteristics of the location itself.

This paper proposes a novel approach that builds upon the domain knowledge of spatial point patterns to model locational information more effectively, which further broadens the scope of location encoding by not specifically transforming location information into vector representations. By focusing on the spatial point pattern effects, rather than each single location that is paired with its corresponding input image, our approach aims to address the limitations of existing methods by incorporating intrinsic spatial properties of locations into the encoding process. This perspective not only enhances the representation of geographic data but also offers a new understanding of spatial relationships and interactions in the context of location encoding. Through this approach, we seek to contribute to the research of GeoAI domain and advance the development of effective and resource-efficient location encoding techniques.

The remainder of the paper is organized as follows. Section 2 describes the GeoAI feature classification workflow and the methods for measuring first- and second-order spatial effects, as well as their integration with the GeoAI models. Section 3 introduces the AI-ready terrain feature data collected for the analysis, along with the experimental design and results. Section 4 concludes our work and discusses future research directions.

## 2 METHOD

In this paper, we thoroughly investigated how geolocation information can help terrain feature recognition under a deep learning framework. We leverage the deep convolutional neural network as the major classifier to decide the categories of terrain features displayed in the given satellite images. The results produced by this classifier are called visional probabilities.

Since geographical images usually come with corresponding geolocations information in the metadata, to support mapping and spatial analysis. To obtain better classification results of terrain features, we exploited such geolocations information and different spatial point pattern analysis (SPPA) methods to statistically produce the different aspects of characteristics of terrain features' distribution, and further decide the likelihood of a terrain feature in a certain location. We named the results produced by these SPPA methods as locational probabilities. An overall framework in figure 1 presents relationships of different building blocks and heterogeneous data sources and how they are integrated together.

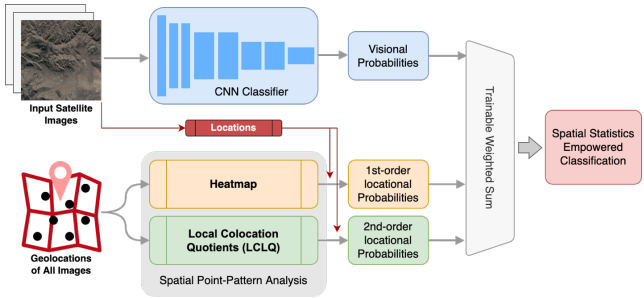

**Figure 1: Overall Framework**

### 2.1 Modeling locational probabilities by the first-order effect of SPPA

In spatial point pattern analysis, the first-order effect focuses on understanding the underlying intensity or density of a spatial point process without considering interactions between points. This approach assumes that the spatial distribution of points can be modeled by a varying intensity function, which describes how the expected number of points per unit area changes across the study area [11]. Essentially, the first-order effect quantifies the spatial intensity or density of points across a study area, reflecting how frequently points occur per unit area. This is often represented using the intensity function, which varies spatially and can be estimated from the observation data. The intensity function $\lambda(x)$ represents the expected number of points per unit area at location $x$. For a spatial point process, this function captures the first-order effect of the point distribution:

$$\lambda(x) = \frac{\text{Expected number of points in a small area around } x}{\text{Area of the small region}}$$

In practice, the intensity function is often estimated using kernel density estimation (KDE). If $\{x_i\}_{i=1}^n$ are the observed point locations, the estimator of the intensity function $\hat{\lambda}(x)$ at location $x$ is given by:

$$\hat{\lambda}(x) = \frac{1}{nh^2} \sum_{i=1}^{n} K\left(\frac{|x - x_i|}{h}\right)$$

Where $n$ is the total number of points, $h$ is the bandwidth parameter, which controls the smoothness of the estimate); $K(\cdot)$ is a kernel function, typically a Gaussian kernel; $|\cdot|$ indicates a Euclidean distance.

When applying the first-order effect to estimate the occurrence probability of different terrain features, we first model the spatial intensity function for each feature type. This involves calculating the density of occurrences for each feature type across the study area using above mentioned intensity function estimator with all observed locations in the training dataset. By estimating how frequently each feature occurs in different locations, we derive intensity maps (aka., heatmaps) per each terrain feature category that reflects the probability of encountering each terrain feature at any given point in the area.

### 2.2 Modeling locational probabilities by the second-order effect of SPPA

The second-order effect extends beyond the first-order intensity function by examining spatial relationships between points, particularly focusing on interactions and clustering within the dataset. While the first-order effect considers how the density of points varies across space, the second-order effect analyzes how the distribution of one type of point influences or is influenced by the distribution of another type. This interaction is crucial for understanding complex spatial structures and dependencies among different features.

The Local Co-location Quotient (LCLQ) [12] is a powerful tool for analyzing these second-order effects. It quantifies the degree to which two types of spatial events occur together more or less frequently than expected by chance. The LCLQ compares the observed density of co-occurrence of two features with their expected density if the features were distributed independently. Specifically, for each location in the study area, the LCLQ is calculated as:

$$LCLQ_{X_i \rightarrow Y} = \frac{N_{X_i \rightarrow Y}}{N_Y/(N-1)}$$

With:

$$N_{X_i \rightarrow Y} = \sum_{j=1(i \neq j)}^{N} \frac{w_{ij} F_Y(j)}{\sum_{j=1(i \neq j)}^{N} w_{ij}}$$

$$w_{ij} = \exp\left(-0.5 \cdot \frac{d_{ij}^2}{h^2}\right)$$

Where $LCLQ_{X_i \rightarrow Y}$ indicates the LCLQ of a point $X_i$ relative to the category $Y$. $F_Y(\cdot)$ is a function returns the binary value indicating whether a point is category $Y$. $w_{ij}$ defined as a Gaussian kernel for density estimation, with $d_{ij}$ indicating the distance between point $i$ and $j$, $h$ denoting the bandwidth. An LCLQ value greater than 1 indicates that the features co-occur more frequently than expected, suggesting a positive spatial association, while a value less than 1 indicates less frequent co-occurrence, suggesting a negative association.

To estimate terrain feature probabilities using the LCLQ, we begin by calculating the LCLQ values for each location $x_i$ in the training dataset. For each location, we derive a vector $V_i = (v_0, v_1, ..., v_C)^T$, where each element $v_j$ represents the LCLQ value corresponding to the $j$-th terrain feature category, and $C$ is the total number of terrain categories. To understand the broader patterns of terrain feature associations across the study area, we compute the global colocation quotient for each terrain feature category by averaging the LCLQ values for each category over all training locations. This results in a global colocation quotient vector for each terrain feature, representing the typical second-order spatial patterns for that feature throughout the study area.

For an arbitrary location in the study area where we want to estimate the probability of different terrain features, we first calculate its LCLQ vector, like the process used for the training locations. To estimate the probabilities of each terrain feature at this location, we compare its LCLQ vector to all global colocation quotient vectors using cosine similarity. A higher cosine similarity indicates a closer match between the local terrain feature associations and the global pattern for a specific feature, suggesting a higher probability of that feature's occurrence at the location.

## 2.3 Deep CNN classifier and probabilities fusion

In our approach to estimating terrain feature probabilities, we first utilize a Deep CNN (DCNN) classifier (e.g., ResNet50) that has been pretrained on ImageNet and further fine-tuned on our specific terrain feature dataset. This fine-tuning process adapts the DCNN model, originally designed for general-purpose image classification tasks, to identify and predict the probabilities of different terrain features from satellite imagery. As a baseline, the classifier outputs a set of predicted probabilities for each feature type, reflecting their likelihood of occurrence at various locations in the study area. To further refine these predictions, a trainable fusion layer implemented as a weighted sum is introduced, which integrates multiple sources of information. This fusion layer combines the probabilities derived from the CNN classifier with those estimated through the abovementioned spatial point pattern analysis methods, specifically the first-order intensity function and the second-order LCLQ. By further fine-tuning the fused model and learning the optimal combination of these different probability sources, the model effectively synthesizes both the feature-specific insights from the CNN and the spatial relationships captured by the point pattern analyses, providing a comprehensive and refined prediction of terrain features.

## 3 EXPERIMENTS AND RESULTS

### 3.1 Data

The dataset used in this work was created by combining satellite imagery, corresponding geographic locations, and category annotations. It is an extension of the GeoImageNet benchmark dataset for terrain feature recognition [13]. The locational data and annotations are sourced from the Geographic Names Information System (GNIS), a comprehensive geographic names database containing over 2 million records of both natural and man-made features across the United States, including Alaska, Hawaii, and other territories. Each GNIS record provides a feature's name, category, and precise geographic coordinates (latitude and longitude) indicating the location of the feature. For our study, we selected a subset of six types of natural features to evaluate our proposed methods: basins, bays, islands, lakes, ridges, and valleys.

We further downloaded corresponding satellite imagery from the National Agriculture Imagery Program (NAIP), based on the location information provided in the GNIS records. NAIP offers ortho aerial imagery with spatial resolutions up to 0.6 meters and updates annually. The imagery used in this work was acquired in January 2023. For each record in our selected subset, we obtained NAIP imagery for a square area with a 6 km length centered on the GNIS-specified location. To ensure compatibility with most deep learning models, the images were further down-sampled to a 6-meter spatial resolution, resulting in an image size of 1000 x 1000 pixels. The category distribution of this dataset, detailing the number of samples for each terrain feature type, is presented in Table 1.

Table 1: Category distribution of the entire dataset

|  | Basin | Bay | Island | Lake | Ridge | Valley |
|---|---|---|---|---|---|---|
| Number of records | 1958 | 5058 | 12558 | 47018 | 12610 | 3667 |

### 3.2 Evaluation of model enhancement by SPPA

We conducted a series of experiments to evaluate the effectiveness of integrating different spatial point pattern statistics into the terrain feature classification model. We specifically examined the impact of incorporating first-order (intensity map) and second-order (LCLQ) spatial effects on the classification accuracy of terrain features using a DCNN model.

The performance metrics for the various configurations are summarized in Table 2, measured with classification accuracy. The configurations tested include the baseline DCNN model, DCNN models enhanced with first-order spatial effect, second-order spatial effect, and a combination of both effects.

Table 2: Comparison of model classification accuracy with and without the integration with spatial point patterns

| Configuration | Validation Accuracy | Testing Accuracy |
|---|---|---|
| DCNN | 0.694 | 0.683 |
| DCNN + 1st-order effect | 0.719 | 0.717 |
| DCNN + 2nd-order effect | 0.703 | 0.690 |
| DCNN + both effects | **0.723** | **0.718** |

Integrating the first-order spatial effect, which captures the overall density and distribution of terrain features, resulted in noticeable performance improvements. Validation accuracy increased by 2.37%, and test accuracy increased by 3.46%. This enhancement highlights the benefit of incorporating spatial density information. When the second-order spatial effect, which models spatial interactions among features, was incorporated, the validation accuracy and the test accuracy increased less than 1% compared with the baseline. While this configuration shows an improvement over the baseline, it is less effective compared to the first-order effect. The relatively modest improvement suggests that spatial interactions, while valuable, may not be as critical as spatial density for our classification tasks. The integration of both first-order and second-order spatial effects resulted in the highest classification accuracy, with a validation accuracy of 0.723 and a testing accuracy of 0.718. The results reveal that integrating different spatial point pattern statistics effectively improves model performance. The first-order effect, which provides information on feature density, leads to substantial improvements in accuracy over the baseline model. Although the second-order is less pronounced, indicating it may play a less crucial role than spatial density in classification task for this dataset, it still contributes to performance gains. This integration enhances the model's ability to capture complex spatial relationships and achieve higher classification accuracy.

## 4 CONCLUSION

This study introduces a strategy of integrating spatial point pattern statistics into terrain feature classification tasks. By employing a GeoAI model enhanced with both first-order and second-order spatial effects, we investigated how incorporating different spatial context impacts the accuracy of terrain feature predictions. The findings of this study demonstrate the effectiveness of integrating spatial point pattern statistics into deep learning models for terrain feature classification. We observed that the second-order effect is less pronounced in improving the model's overall prediction accuracy. We speculate that it might be caused by the inherent spatial patterns of the dataset or the modeling effectiveness of the second-order effect. In the future, we will focus on further refining these point pattern analysis methods, exploring additional spatial effects, and testing across diverse GeoAI models and datasets (e.g., [14]) to validate the method's generalizability and robustness.

## ACKNOWLEDGMENTS

This work is supported in part by the National Science Foundation under awards 1853864, 2120943, 2230034, as well as Google.org's Impact Challenge for Climate Innovation Program.